\documentclass[letterpaper,10pt]{article}
\usepackage{graphicx}
\usepackage{setspace}
\usepackage{float}
\usepackage{amssymb}

\newcommand{\ignore}[1]{}

\newcommand{\nscom}[1]{\footnote{!!NS!! #1}}
\newcommand{\glcom}[1]{\footnote{!!GL!! #1}}
\newcommand{\dhcom}[1]{\footnote{!!DH!! #1}}
\newcommand{\hmcom}[1]{\footnote{!!HM!! #1}}

\renewcommand{\nscom}[1]{}
\renewcommand{\hmcom}[1]{}
\renewcommand{\glcom}[1]{}
\renewcommand{\dhcom}[1]{}

\title{Application of Support Vector Regression to Interpolation of Sparse Shock Physics Data Sets}

\author{
Nikita A. Sakhanenko{\small $^{1,2}$}~~~ George F. Luger\\ 
{\small \{sanik,luger\}@cs.unm.edu}\\
{\small $^1$Computer Science Department, University of New Mexico}\\
\\
Hanna E. Makaruk~~~  David B. Holtkamp\\
{\small \{hanna\_m,dholtkamp\}@lanl.gov}\\
{\small $^2$Physics Division, Los Alamos National Laboratory}
}

\date{}

\begin{document}

\maketitle

\begin{abstract}

Shock physics experiments are often complicated and expensive. As a result, researchers 
are unable to conduct as many experiments as they would like -- leading to sparse data sets. 
In this paper, Support Vector Machines for regression are 
applied to velocimetry data sets for shock damaged and melted tin metal. Some success at 
interpolating between data sets is achieved. Implications for future work are discussed.

\end{abstract}

\section{Introduction}

Experimental physics, along with many other fields in applied and basic research, 
uses experiments, physical tests, and observations to gain insight 
into various phenomena and to validate hypotheses and models. Shock physics is a field 
that explores the response of materials to the extremes of pressure, deformation, 
and temperature which are present when shock waves interact with those materials \cite{Zeldovich02}.  
High explosive (HE) or propellant guns are often used to generate these strong shock 
waves. Many different diagnostic approaches have been used to probe these phenomena \cite{Isbell05}.  

Because of the energetic nature of the shock wave drive, often a large amount of 
experimental equipment is destroyed during the test. Like many other applied sciences, 
the cost and complexity of repeating a significant number of experiments -- or 
conducting a systematic study of some physical property as a function of another -- 
are simply too costly to conduct to the degree of completeness and detail that a 
researcher might desire. Often a researcher is left with a sparse data set -- one 
that numbers too few experiments or samples a systematic variation with too few points.

The present work applies Support Vector Machine techniques to the analysis 
of surface velocimetry data taken from HE shocked tin samples using a laser velocity 
interferometer called a VISAR \cite{BarkerHollenback72,BarkerSchuler74,Hemsing79}.  
These experiments have been described elsewhere in detail \cite{Holtkamp03}. 
For the purposes of this paper, it is sufficient that the VISAR velocimetry data 
presented here describe the response of the free surface of the metal coupon to the shock 
loading and release from the HE generated shock wave. The time dependence of the magnitude 
of the velocity can be analyzed to provide information on the yield strength of the 
material, and the thickness of the leading damage layer that may separate from the 
bulk material during the shock/release of the sample.\nscom{This para should be 
more clear in CS version.}

In section \ref{sec:problem} we describe the problem and
include more details on the VISAR system (section \ref{sec:visar_details}).
In section \ref{sec:approach} the Support Vector Machine technique is 
presented, and its applicability for our problem is discussed. In section
\ref{sec:results} we evaluate the results achieved. Finally, the paper ends
with a discussion of related work and conclusions.

\section{Problem definition}
\label{sec:problem}

\subsection{Metal melting on release}
\label{sec:experiment}

This paper is based on the data obtained from experiments when metal samples are
damaged/melted after a high explosive detonation with a single point ignition. 
\begin{figure}[h]
  \centering
  \includegraphics[height=.3\textheight]{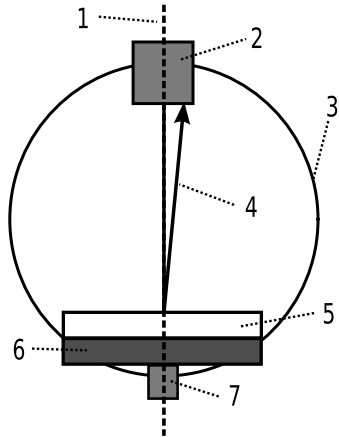}
  \caption{Schematic representation of the experiment setup: (1) the axis of symmetry
of an experiment, (2) VISAR probe, (3) a view area captured by PRAD imagery, (4) reflected
laser light, (5) a metal coupon, (6) high explosive coupon, (7) detonator.}
  \label{experiment}
\end{figure}
A schematic view of the experiment setup is shown in figure \ref{experiment}.
A cylindrically shaped metal coupon is positioned ontop of $0.5$ inch
thick high explosive (HE) disk. Both the metal and HE coupons are $2$ inches
in diameter. A point detonator is glued to the center
of the HE disc in order to perform a single point ignition symmetrically.
Note that all of the components of the experiment setup have a common 
axis of symmetry. During an experiment a VISAR probe, located on the
axis above the metal sample, transmits a laser beam, and the velocity
of the top surface of the metal is infered from the Doppler shifted light
reflected from the coupon (see section \ref{sec:visar_details}
for more details). The time series of the velocity measured through
out an experiment constitutes the VISAR velocimetry.

In the same experiment a proton beam is shot perpendicular to the
experiment's axis. By focussing the beam, a Proton Radiography (PRAD)
image of the current experiment state is obtained. This is somewhat
similar to X-ray imagery, although Proton Radiography can produce up
to $20$ or $30$ images in a single experiment with an image exposure
time of $< 50$ns. This paper
is devoted to VISAR velocimetry data analysis, while discussion
about Proton Radiography imagery analysis may be found elsewhere 
\cite{Holtkamp03}.

There are two parameters that vary between different 
experiments: the metal type of the sample and the thickness of the
coupon. By changing the thickness of the metal coupon and the type of metal 
in the initial setup of an experiment, experimentalists
attempt to see the changes in physical processes across the set of
experiments. For simplicity, only the experiments on tin samples are
described in this paper.

\subsection{VISAR data}
\label{sec:visar_details}

\nscom{Note that CS version of this
section has to be smaller and simpler.}

A Velocity Interferometer System for Any Reflector (VISAR) is
a system designed to measure the Doppler shift of a laser beam
reflected from the moving surface under consideration so as to
capture changes of the velocity of the surface. The VISAR system
is able to detect very small velocity changes (few meters per second).
Moreover, it is able to measure even
the velocity changes of a diffusely reflecting surface.

\begin{figure}[h]
  \centering
  \includegraphics[height=.2\textheight]{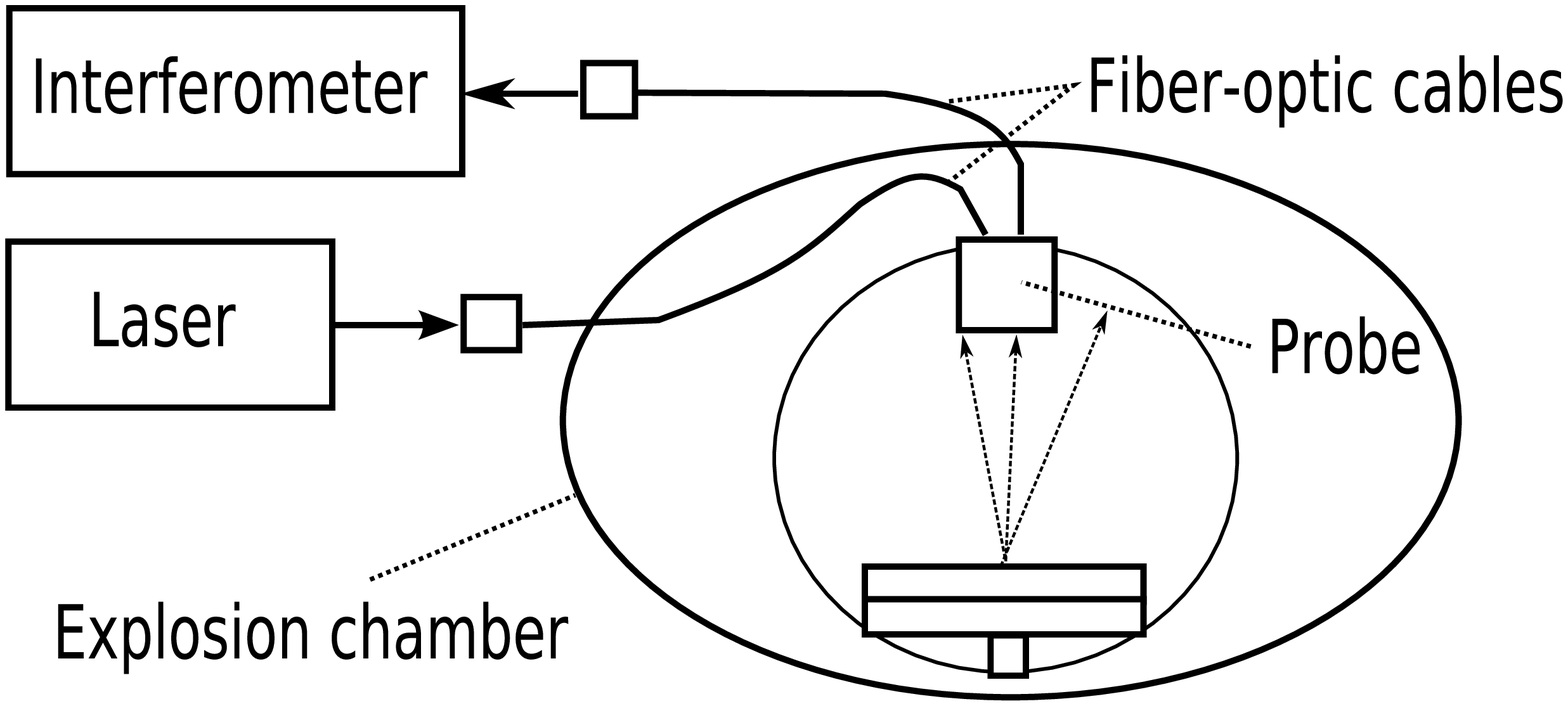}
  \caption{Schematic view of a VISAR system.}
  \label{visar1}
\end{figure}

A VISAR system consists of lasers, optical elements, detectors, and other 
components as shown in figure \ref{visar1}.
The light is delivered from the laser via optical fiber to the probe and
is focussed in such a way that some of the light reflects from the moving
surface back to the probe. The reflected laser light is transmitted
to the interferometer. Note that since the reflected light is Doppler
shifted, one can extract the velocity of the moving surface from the 
wavelength change of the light. The interferometer is
able to identify the increase or decrease of the wavelength of a beam.

\begin{figure}[h]
  \centering
  \includegraphics[height=.3\textheight]{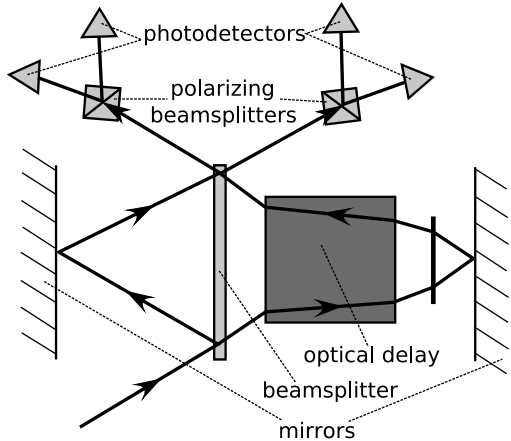}
  \caption{Schematic view of the VISAR interferometer. Note that two beams
obtained after splitting the initial one travel different optical distances.}
  \label{visar2}
\end{figure}

The captured Doppler-shifted light, the frequency of which is different from
that of the initial beam, is transmitted into the interferometer 
depicted in figure \ref{visar2}, where the beam is split into two.
Using optics, two beams travel different optical distances.
By adjusting the length of the paths of the
beams, the beams are made to interfere with each other
before reaching the photodetectors. Finally,
the information is extracted from a VISAR system by measuring the intensity signals
from the photodetectors. For more details on the VISAR system consult 
\cite{BarkerHollenback72,BarkerSchuler74,Hemsing79}.

This method, widely used in the experiments similar to the one described in
section \ref{sec:experiment}, is reasonably reliable. For instance,
the measurements obtained using a VISAR system are in agreement with the results
obtained by Makaruk et al. \cite{Makaruk06} after positions of different fragments
visible on a PRAD image were measured and their corresponding velocity was computed.
Since this method of information extraction is independent of VISAR, it 
additionally validates VISAR results.

\subsection{Filling gaps of VISAR data}
\label{sec:filling}

The problem considered in this paper, given a limited number of experiments
that are difficult and costly to perform, is to estimate the measurement values for
the missing experiments, or the experiments, whose data recordings were
not successful. This problem is also strongly related to the one of
identifying ``outlier'' experiments, i.e., those experiments that for some 
reason went wrong. The data estimation methods can show which experimental 
data do not fit with other ``good'' experiments.

The task of increasing the informational output of VISAR data is important,
due to the limited number of experiments, their difficult implementation, and high cost.
Physicists, who attempt to explain all the phenomena of
these experiments, can gain better physical insight from the combination of the VISAR 
data and the estimations than from the experimental data above.
Another important application of velocity estimations is for comparison 
with various kinds of hydrocode models generated by large programs\footnote{A hydrocode simulation of an
experiment is based on a set of physical equations defining the relevant physical
laws. The simulation starts from the same initial conditions as the real
experiment. The simulations are performed in 2-dimensional or 3-dimensional
space, depending on the type of a hydrocode. Note that these simulations are
frequently called numerical experiments.} that simulate shock or other
hard/impossible to perform experiments. The PRAD data, and the other
type of information collected during these experiments, can also be
supported and even improved by extending the velocity estimations.

\ignore{
The above descriptions show how important it is to compute good VISAR estimations
not only for VISAR data analysis, but also for understanding of the physical system
as a whole.}

\section{Our approach}
\label{sec:approach}

\subsection{Equivalent problem}

Recall that each VISAR data point is a triple $\langle t,w,v \rangle$, where $t$ is 
the time when the measurement was recorded, $w$ is the thickness of a sample, 
and $v$ is a measured velocity. One can
see that these data points lie on a $2$-dimensional surface in the $3$-dimensional 
space. Hence the problem identified in section \ref{sec:filling} can be transformed into the task of 
reconstructing the $2$-dimensional surface from the given VISAR data.

In other words, the problem is to find a regression of velocity on the time and 
thickness of a sample. Formally, given three random variables that map a probability
space $(\Omega,A,P)$ into a measure space $(\Gamma,S)$, velocity, time, and thickness
$V,T,W:(\Omega,A,P)\rightarrow(\Gamma,S)$, the problem is to estimate coefficients 
$\lambda \in \Lambda$ such that the error $e = V-\rho(T,W;\lambda)$ is small. Here
$\rho$ is a regression function that is $\rho:\Gamma^2\times\Lambda\rightarrow \Gamma$,
where $\Lambda \subseteq \Gamma$ is some set of coefficients. 
In the case of the problem considered
in this article, $\Gamma=\mathbb{R}$. Note that variables $T$ and $W$ are the two factors of
a regression, and $V$ is an observation.

\subsection{Velocity surface reconstruction using Support Vector Regression}

\hmcom{Physics version needs more general 
explanation what is SVM, where it was successfully applied, and what physicist 
may expect from it.}

The Support Vector Machine (SVM) uses {\em supervised learning} to 
estimate a functional input/output relationship from a training data set. 
Formally, given the training data set of $k$ points
$\{ \langle x_i,y_i \rangle | x_i \in X, y_i \in Y, i=1 \ldots k \}$, 
that is independently and randomly generated by some unknown function $f$ 
for each data point, the Support Vector Machine method finds an approximation of the 
function assuming $f$ is of the form 
\begin{equation}
f(x)=w\cdot \phi(x)+b,
\label{eq1}
\end{equation}
where $\phi$ is a 
nonlinear mapping $\phi:X \rightarrow H$, $b \in Y$, $w \in H$. Here $X \subseteq R^n$
is an input space, $Y \subseteq R$ is an output space, and $H$ is a high-dimensional feature
space. The coefficients $w$ and $b$ are found by minimizing the {\em regularized risk} 
\cite{Scholkopf01} $R=\sum^k_{i=1}Loss(f(x_i),y_i)+\lambda \parallel w \parallel^2$ that is
an empirical risk, defined via a loss function, complemented with a regularization term. 
In this paper we use an $\varepsilon$-intensive Loss Function \cite{Vapnik99} 
defined as 
$$
Loss(f(x),y)=\left\{
\begin{array}{ll}
|f(x)-y|-\varepsilon & if |f(x)-y| \geq \varepsilon\\
0 & otherwise
\end{array}\right.
.
$$
Note also that the Support Vector Machine is a method involving kernels. Recall that 
the kernel of an
arbitrary function $g:X \rightarrow Y$ is an equivalence relation on $X$ defined as
$$
ker(g)=\{(x_1,x_2)|x_1,x_2\in X,g(x_1)=g(x_2)\}\subseteq X\times X.
$$

Originally, the SVM technique was applied to classification problems, 
in which the algorithm 
finds the maximum-margin hyperplane in the transformed feature space $H$ that separates 
the data into two classes. The result of an SVM used for regression estimation (Support Vector
Regression, SVR) is a model that depends only on a subset of training data, because
the loss function used during the modeling omits the training data points inside the 
$\varepsilon$-tube (points that are close to the model prediction).

We selected the SVM approach for this problem because of the attractive 
features pointed out by Shawe-Taylor and Cristianini 
\cite{ShaweTaylorCristianini04}. 
One of these features is the good generalization performance 
which an SVM achieves by using a unique
principle of structural risk minimization \cite{Vapnik79}. In addition, SVM training is
equivalent to solving a linearly constrained quadratic programming problem that has a unique
and globally optimal solution, hence there is no need to worry about local minima.\nscom{
In physics version last 2 sentences should be simpler and clearer.} 
A solution found by SVM depends only on a subset of training data points, 
called {\em support vectors}, making the representation of the solution sparse. 

Finally, since the SVM method involves kernels, it allows us to deal with 
arbitrary large feature spaces without having to compute explicitly the mapping $\phi$ 
from the data space to the feature space, hence avoiding the need to compute 
the product $w \cdot \phi(x)$ of (\ref{eq1}). In other words, a linear algorithm 
that uses only dot products can be transformed 
by replacing dot products with a kernel function. The resulting algorithm becomes
non-linear, although it is still linear in the range of the mapping $\phi$. We do not
need to compute $\phi$ explicitly, because of the application of kernels.
This algorithm transformation from the linear to non-linear form is known as a so-called 
{\em kernel trick} \cite{Aizerman64}. 

On the other hand, since the available data are the VISAR 
measurements that capture some characteristics of
the unknown function, and each data point is represented by 
several features, the data is suitable for the application of
supervised learning methods, such as SVR. A velocity of each data point 
is a target value for SVR, whereas the thickness and time are feature values.
\vspace{-40pt}
\begin{figure}[h]
  \centering
  \includegraphics[height=.4\textheight]{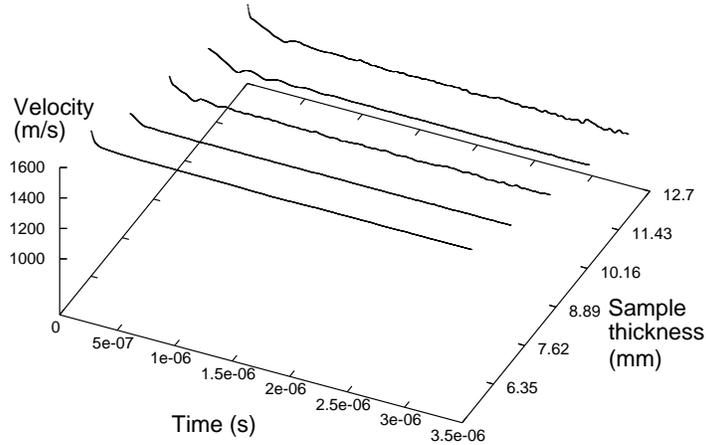}
  \caption{Available VISAR data set}
  \label{data3d}
\end{figure}

In figure \ref{data3d} the VISAR data set is depicted. It is important 
to note that the data is significantly stretched along the time dimension. 
This happens, because the whole dataset is comprised of a number of time 
series corresponding to a set of measured experiments. During each experiment, 
the VISAR readings were recorded every $2$ns for as long as $6000$ time steps. 
However, for some of the experiments the VISAR system finished recording useful
information earlier than for others. The data were cut by the shortest
sequence ($1656$ time steps), since it has been identified experimentally that 
SVM performs better on the aligned data. On the other hand, if we consider
VISAR measurements across the thickness dimension, the data covers the
thicknesses starting from $0.25$ inch up to $0.5$ inch with $0.0625$ inch increase. 
In total $5$ time sequences of $1656$ points comprise the data used by the SVM method.

\begin{figure}[h]
  \centering
  \includegraphics[height=.4\textheight]{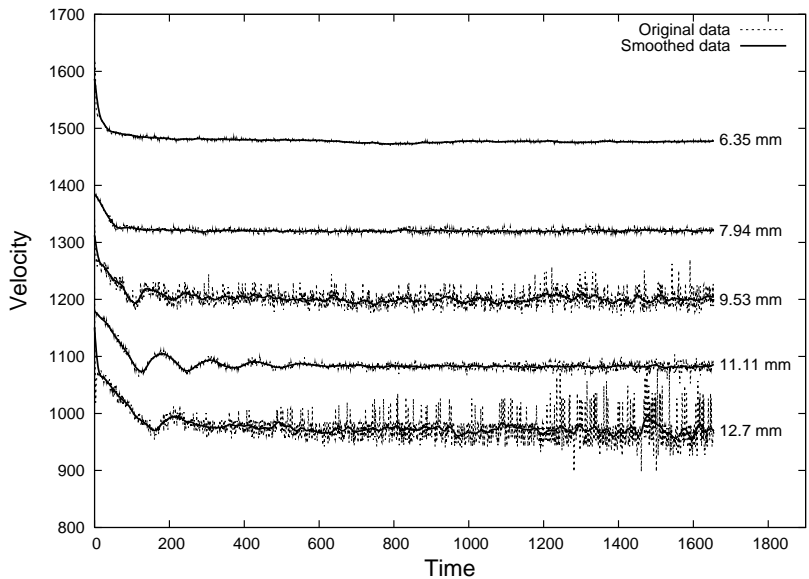}
  \caption{The projection of the VISAR data set and its smoothed version}
  \label{smoothed}
\end{figure}
Figure \ref{smoothed} presents the complete data set 
projected on the $Time \times Velocity$ plane.
The original data is represented by dotted lines, and its 
smoothed with a sliding triangular
window version is depicted with solid lines. Note that the 
amount of the time steps, where each step
is equal to $2$ns, is shown on the abscissa.

In order to identify the best application of the SVM method to the VISAR data,
we use standard $k$-fold cross-validation. The data is divided into $k$ parts,
out of which $k-1$ parts are used for training the learning machine, and the
last part is used for its validation. The process is repeated $k$ times using
each part of the partitioning precisely once for validation.

\section{Evaluation/Results}
\label{sec:results}

There are several factors affecting the quality of the resulting regression. The error of 
VISAR data and the errors occurring during the data preprocessing affect the accuracy
of the reconstructed surface the most. It is generally agreed that a VISAR system 
measures the velocity values with an absolute accuracy of 3-5$\%$. This is
an approximate error calculated from differences between repeated experiments.
Although the number of repeated experiments was too
small to allow a more robust statistical analysis, this level of
uncertainty is in the range of values generally agreed on by VISAR
experimenters \cite{BarkerHollenback72,BarkerSchuler74,Hemsing79}.
This error together with the noise transfers 
into the regression result. In addition, since the ignition time 
(the start of the experiment)
was different for different experiments, the data has to be time-aligned so as to make each
time series start exactly from the moment of the detonation. This introduces another 
potential error into the regression.

The accuracy of the reconstructed surface is also affected by the specific features of
VISAR data. The length of each of the time series produced by the VISAR system during 
different experiments always differs. We have observed that the SVM performs better on the data
combined from the time series of the same length than from those of different length. Hence,
the length of the data was aligned. In addition, each data point of three elements (velocity, 
time, and thickness) has order $10^3$, $10^{-6}$, and $1$. This is why it is important
to scale the data to improve the performance of the SVM.

Unfortunately, the application of SVR directly to the set of smoothed and aligned data 
yields overfitted 
results, because the data step in the time direction is much smaller than
the step in the other directions, and hence for any chosen data range there are more 
data points along the time axis
than along the thickness axis. The overfitting problem is solved by scaling the data in
such a way that the distance between two neighbor points along any axis is equal to $1$.

Using nonlinear kernels achieves better performance, when the dynamics
of an experiment are non-linear. It is known that Gaussian Radial Basis Function (RBF) kernels
perform well under general smoothness assumption \cite{Smola98}, hence
a Gaussian RBF 
$$
k(x,y)=e^{- \gamma \parallel x-y \parallel^2}
$$ 
has been chosen
as a kernel for the reconstruction. Additionally, it has been experimentally determined
that SVM techniques with simpler kernels, such as polynomial, take longer time to 
train and return non-satisfactory results.

The performance of the SVR with RBF kernel is directly affected by three parameters, the radius
$\gamma$ of RBF, the upper bound $C$ on the Lagrangian multipliers
(also called a regularization constant or a capacity factor), 
and the size $\varepsilon$ of the $\varepsilon$-tube (also called an error-insensitive zone or 
an $\varepsilon$-margin). Note that $\varepsilon$ determines the accuracy of the regression,
namely the amount by which a point from a training set is allowed to diverge from the
regression. $k$-fold cross-validation is performed in order to determine the optimal 
parameters' values under which SVR produces 
the best approximation of the surface. An $l^2$ error
is computed for each parameter 
instantiation after finishing the cross-validation. 
\begin{figure}[h]
  \centering
  \includegraphics[height=.37\textheight]{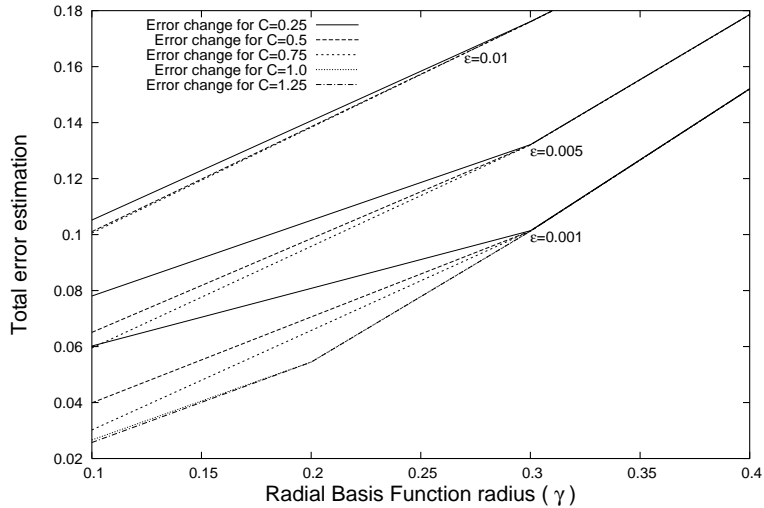}
  \caption{Error changes depending on different model parameters}
  \label{gerror}
\end{figure}
Figure \ref{gerror} demonstrates
how the error changes depending on the values of the SVR parameters.

\begin{figure}[h]
  \centering
  \includegraphics[height=.4\textheight]{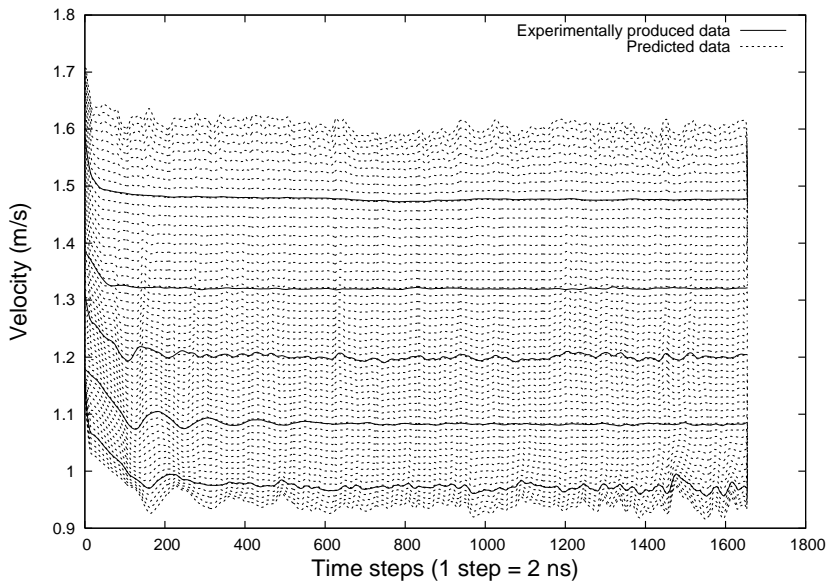}
  \caption{SVM prediction results: dotted lines represent the prediction of the time series
for the thicknesses between those that are produced experimentally (the solid lines).}
  \label{surface}
\end{figure}

It can be seen in figure \ref{gerror} that the error increases as the radius $\gamma$ 
goes up. The error also increases when $\varepsilon$ becomes bigger. One can also see 
that the change of $C$ affects the error the most when $\gamma$ is the smallest, and 
the influence of $C$ on the error decreases as $\gamma$ goes up, becoming insignificant
when $\gamma$ exceeds $0.3$. In the same time, given a small $\gamma$, parameter $C$
affects the error more as the $\varepsilon$ decreases. The error analysis suggests that
when the tuple $\langle \gamma, C, \varepsilon \rangle$ is around 
$\langle 0.1, 0.75-1.0, 0.001 \rangle$, the error is the smallest. This error 
analysis produces a range of suboptimal values for the parameters.
The expert knowledge is used
in order to identify the final model that returns the most
accurate velocity surface, shown in figure \ref{surface}.

Once the surface is found, it is possible to obtain a velocity value for any 
$\langle time,thickness \rangle$ pair. Assuming the surface is accurate enough, 
the failed VISAR data that deviates considerably from the surface can be 
identified. The surface provides significantly more information about
velocity changes across the thickness dimension than VISAR readings alone. 
It can also provide velocity time series for an experiment, in which only 
PRAD data were measured successfully, improving the quality of the analysis 
for this experiment, and, consequently, increasing the
understanding of the whole physical system.

It should be noted that in this paper we used an implementation of the 
SVM technique called {\em SVM-light}. 
For more information about its implementation details see \cite{Joachims99}.

\section{Related Work}
\label{sec:related_work}

\hmcom{For physics type of journal 
the examples of successful use in physics are the most important.}

Even though one can encounter many different applications of SVM in various fields, most of
them are used for classification. Support vector classification methods are successfully used in
fields such as image analysis and pattern recognition, speech recognition, bioinformatics,
e-learning, and others. Compared to the classification case, support vector regression as
a variation of the SVM technique has not been used in many problems outside AI. Hence,
it is not a surprise that SVM methods were never before applied to VISAR data, nor in a 
broader sence, to experimental physics environments. Vannerem et al. in
\cite{Vannerem99} attempted to test SVM in this environment by using support vector classifiers
in the analysis of simulated high energy physics data. In \cite{Cai03} Cai et al. presents 
another example of SVM application in the analysis of physics data. In that paper the
support vector machine is used to classify sonar signals.

As noted above, SVM for regression (as opposed to SVM classification) is rarely applied 
in physics. There are, however, several
successful examples of the support vector regression application. In \cite{Dibike01} 
Dibike et al. introduced the regression type of the SVM technique to the civil engineering 
community and showed that SVM can be successfully applied to the problem of stream
flow data estimation based on records of rainfall and other climatic data. By using three
types of kernels, Polynomial, RBF, and Neural Network, and choosing the best values for
SVM free parameters via trial and error, the authors point out that the SVM with the 
RBF kernel performs the best. Finally, this research is the first attempt to apply 
support vector regression in data analysis of VISAR measurements obtained
from experiments on shock melted and damaged metal.

\section{Conclusions and Future Work}
\label{sec:conclusion}

In this paper we described the problem of VISAR data analysis in which we attempted to
estimate the data between the points measured by VISAR. The Support Vector Regression
method was used to reconstruct a 2-dimensional velocity surface in 
$Time \times Thickness \times Velocity$ data space resulting in a successful estimation.
The SVR free parameters were obtained from a grid search as well as using the expert
knowledge.

The velocity surface provides considerably more information about the velocity behavior
as a function of time and thickness than experimentally produced VISAR measurements 
alone. This may significantly improve the scientific value of VISAR data into other
areas of analysis of shock physics experiments, such as PRAD imagery analysis and
hydrocode simulations.

On the other hand, support vector regression does not require a vast amount of data
for producing good velocity estimations. This is very helpful because of the high cost
and complexity of experiments, and limited amount of available data.

In addition, the estimated velocity surface can help to identify experiments-outliers:
those experiments that for some reason went wrong. The data obtained from these 
experiments will be significantly different than those suggested by the velocity
surface.

There are several directions in which this work might advance. One of these is to 
investigate the possibility of using a custom kernel instead of a standard 
Gaussian. Intuitively,
an elliptical kernel that accounts for high density of the data in one direction and 
sparsity in all other directions may improve the results of support vector regression.

During regression performed by the support vector machine method, we need to identify
optimal values for SVM free parameters. In this paper a grid search and the expert 
knowledge are used (see section \ref{sec:results}), essentially leading to suboptimal 
parameter values. Investigation of deriving an online learning algorithm
for SVM parameter fitting specific to the VISAR data might be another direction of 
further research.

In addition, note that an SVM system used for regression outputs a point estimate. 
However, most of the time we wish to capture uncertainty in the prediction, hence
estimating the conditional distribution of the target values given feature values
is more attractive. There is a number of different extensions to the SVM technique
and hybrids of SVM with Bayesian methods, such as {\em relevance} vector machines and
Bayesian SVM, that use probabilistic approaches. Exploring these methods could 
give significantly more information about the underlying data.

\section{Acknowledgements}

The authors are thankful to Joysree Aubrey for the numerous long and thought-provoking 
discussions. Special thanks to Brendt Wohlberg for providing ideas about SVM applications.
This work was supported by the Department of Energy under the ADAPT program.

\end{document}